# LLM-Gomoku: A Large Language Model-Based System for Strategic Gomoku with Self-Play and Reinforcement Learning

Hui Wang

Peking University

wanghui22@stu.pku.edu.cn

## Abstract

In recent years, large language models (LLMs) have shown significant advancements in natural language processing (NLP), with strong capabilities in generation, comprehension, and reasoning. These models have found applications in education, intelligent decision-making, and gaming. However, effectively utilizing LLMs for strategic planning and decision-making in the game of Gomoku remains a challenge. This study aims to develop a Gomoku AI system based on LLMs, simulating the human learning process of playing chess. The system is designed to understand and apply Gomoku strategies and logic to make rational decisions. The research methods include enabling the model to "read the board," "understand the rules," "select strategies," and "evaluate positions," while enhancing its abilities through self-play and reinforcement learning. The results demonstrate that this approach significantly improves the selection of move positions, resolves the issue of generating illegal positions, and reduces process time through parallel position evaluation. After extensive self-play training, the model's Gomoku-playing capabilities have been notably enhanced. [1]

## 1 Introduction

In recent years, large language models (LLMs) have made remarkable progress in the field of natural language processing (NLP), demonstrating their prowess in generation, comprehension, and reasoning across various tasks (Zhao et al., 2025). For instance, models like GPT4 (OpenAI et al., 2024), Gemini-2.0 (Team et al., 2024) , DeepSeek-V3 (DeepSeek-AI et al., 2025), Llama3 (Grattafiori et al., 2024) , Qwen2.5 (Qwen et al., 2025) et al. through pretraining and fine-tuning, can handle complex language tasks and deliver outstanding performance even with few or no samples. Moreover, LLMs have been widely applied in education (Wang et al., 2024), intelligent decision-making (Li et al., 2025), and gaming (Hu et al., 2024), showcasing their broad potential. In education, LLMs can act as intelligent agents to support personalized learning and enhance students' reading and writing skills. In intelligent decision-making, they can analyze vast amounts of data to provide valuable suggestions and solutions for decision-makers. In gaming, LLMs can generate engaging content and interactive experiences.

However, despite their success in many tasks, applying LLMs to specific complex tasks remains challenging (Kaddour et al., 2023). For example, in the strategic game of Gomoku, effectively leveraging LLMs for rational strategy planning and decision-making is an area worth exploring (Topsakal et al., 2024). Gomoku, a classic strategic board game, is beloved for its simple rules and profound strategic depth (Piazzo et al., 2021). Traditional search algorithms like exhaustive search perform poorly under limited computational resources, while machine learning-based methods, though powerful, suffer from low training and prediction efficiency (Xie et al., 2018). Therefore, integrating the strengths of LLMs with deep learning and reinforcement learning to design an efficient and accurate Gomoku AI is an important direction for current research.

This study aims to develop a Gomoku AI system based on large language models enabling it to understand and apply Gomoku strategies and logic like human players, thereby making rational and effective decisions during games. The core of this research lies in simulating the human learning process of playing chess, with the key focus being "how to select appropriate strategies and analytical logic based on the current game situation."

---

[1] https://github.com/stranger47MCI/LLM-Gomuku

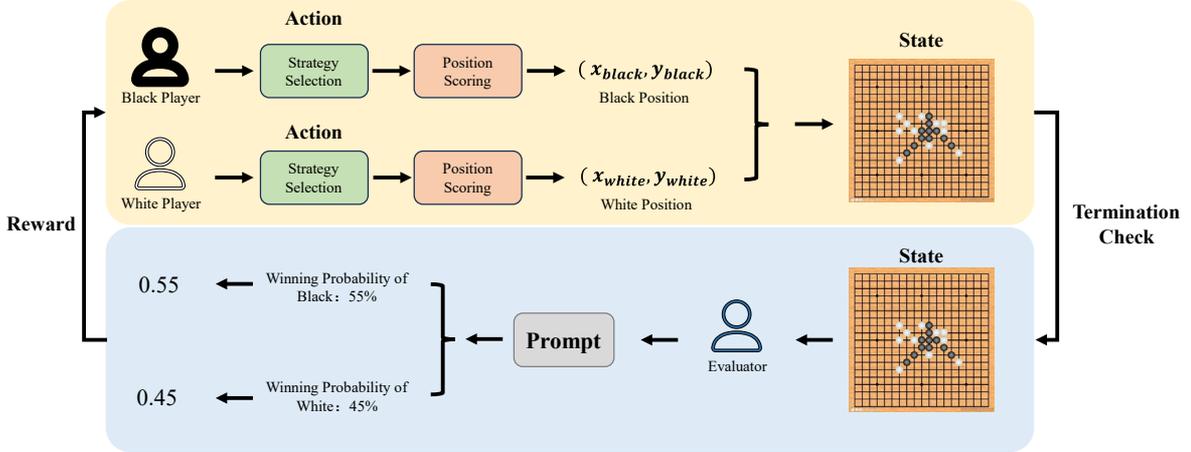

Figure 1: Flowchart of the Overall Process of LLM-Gomoku.

Firstly, the model is made to "read the board" by inputting the board information, enabling it to accurately identify the current state of the board, including the positions of both its own and the opponent's pieces, as well as the boundaries of the board. This provides the foundation for subsequent decision-making. Secondly, the model is taught to "understand the rules" by inputting the basic rules of Gomoku, such as the placement of pieces and the conditions for winning, ensuring that the model operates according to the rules during the game. Next, the model is made to "learn chess strategies" by inputting a variety of different strategies, such as the first-move layout, defensive counterattack, and consecutive attack, allowing the model to flexibly apply corresponding strategies according to different game situations. Then, the model is made to "analyze the game situation" by inputting various analytical logics, such as judging the situation, speculating the opponent's intentions, and predicting the next move, to cultivate the model's ability to analyze the game situation. After that, the model is made to "try playing chess" by engaging in self-play (Zhang et al., 2025), accumulating experience and optimizing decisions through continuous game practice. Finally, the model's chess-playing level and strategy application ability are gradually improved through "practical combat enhancement," using reinforcement learning methods (Ghasemi and Ebrahimi, 2024) to let the model continuously learn and progress in games with human players or other intelligent systems, ultimately achieving a high level of Gomoku playing ability.

## 2 Relate Work

The application of Large Language Models to gaming has attracted significant research interest following recent model advancements. Parallel lines of investigation have emerged: one focusing on basic competency assessment through simple deterministic games like Tic-Tac-Toe (Topsakal et al., 2024), another developing comprehensive frameworks for evaluating strategic reasoning across diverse game-theoretic environments (Duan et al., 2024). These studies collectively identify a consistent performance dichotomy - LLMs exhibit particular difficulty with deterministic, complete-information games compared to probabilistic scenarios, a finding especially relevant for complex strategy games like Gomoku and central to our investigation of reinforcement learning approaches for decision-making enhancement.

Theoretical examinations from game-theoretic perspectives have established foundational insights into LLMs' strategic rationality (Fan et al., 2023). Complementary empirical work has systematically documented the critical role of prompt engineering in shaping LLMs' game strategy processing and spatial reasoning capabilities (Liga and Pasetto, 2023), directly influencing our Gomoku prompt design methodology. Despite demonstrable progress in recent multi-game evaluations, a persistent performance gap remains between LLMs and human-level strategic reasoning (Costarelli et al., 2024). This limitation, highlighted in broader surveys of LLM-based gaming agents, points to the necessity of developing more sophisticated frameworks for in-game strategy learning and decision-making processes (Hu et al., 2024).

Our study builds on these findings, focusing on enhancing LLM performance in the complex strategy game of Gomoku through reinforcement learning. We aim to contribute new insights into leveraging LLMs for strategic game play and expand their potential applications.

## 3 Methods

In this section, we first introduce the overall logic of the code implementation, which is primarily divided into five aspects: 1) Prompt design, 2) strategy and analytical logic selection, 3) local position evaluation, 4) self-play, and 5) reward model and reinforcement learning. We then elaborate on the specific details of each component in turn.

### 3.1 Prompt Design

To enable large language models (LLMs) to more accurately simulate the decision-making process of human players in Gomoku, we designed a general-purpose prompt template (Schulhoff et al., 2025). This template aims to replicate the thought process of human players during a game by incorporating key elements such as the current state of the board, the basic rules of Gomoku, chess strategies, and analytical logic. The specific prompt template is shown in Figure 2. Through this approach, we hope that the language model can better understand and simulate the decision-making logic of human players, thereby making more rational and efficient move choices in Gomoku games.

**Prompt-1**

"You are an outstanding Gomoku player. These are the basic rules of Gomoku: {rule}. Later, I will need your help to play Gomoku.

I will be playing on a board of size { self.board_size}*{self.board_size}, and I will provide an array representing the state of the board from left to right, line by line, where { self.player_id } indicates my pieces, { -self.player_id } indicates my opponent's pieces, and 0 indicates empty spaces.

Please use the {think} strategy to analyze the behavior of both me and my opponent, and use the {relationship} to understand the current situation. Please ensure not to place a piece on an already occupied position. Remember, only choose the one you consider the best from the { zero_position } positions!"

Figure 2: Prompt Template for Move Position Consideration.

### 3.2 Strategy and Analytical Logic Selection

In Gomoku games, human players often rely on chess strategies and game records to select moves with higher chances of winning, and they need to choose appropriate analytical methods to determine how to apply specific strategies. Inspired by this, our study aims to simulate the thought process of human players by collecting and learning common Gomoku strategies and analytical logics, so as to accurately select suitable strategies and logics for thinking when facing a game situation.

Specifically, we have collected 52 common chess strategies, covering four key parts: basic tactics, defensive strategies, offensive strategies, and opening methods. At the same time, we have also collected 9 types of analytical logics, such as causal relationships, conditional relationships, and comparative relationships. During the thinking process, the large language model will select one strategy and one logic from the collected ones for in-depth thinking. To further improve the accuracy of the large language model's analysis, we will also incorporate the rules of Gomoku and the specific information of the current game situation into the thinking process prompt and output the most reasonable

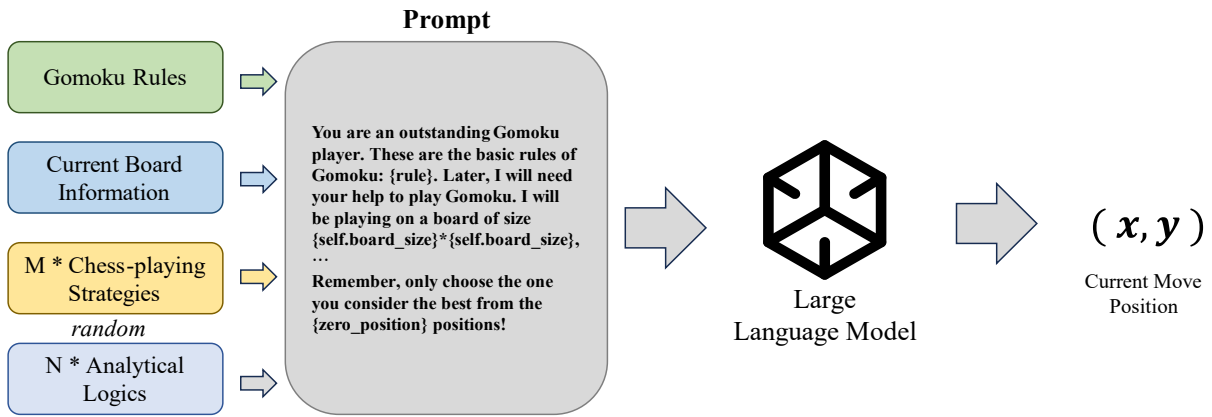

Figure 3: Flowchart of the Overall Process of Integrating Strategy and Analytical Logic Selection into the Thinking Process and Outputting the Move Position

move location. The specific process is shown in Figure 3.

## 3.3 Local Position Evaluation

During Gomoku games, large language models often encounter the issue of selecting illegal move positions, especially placing pieces on already occupied spots. To address this problem, common methods include repeatedly thinking until a legal position is found or incorporating illegal position information into the prompt in advance.

Experimental results show that while these two methods can effectively alleviate the issue of selecting illegal positions in the early stages of the game, they still fail to completely overcome this problem as the number of pieces on the board increases. This leads to infinite loops in the thinking process and program crashes.

To tackle this challenge, we have carefully designed a local position evaluation method. This method evaluates each legal position among the candidate move positions and their local neighbors, precisely selecting the highest-scoring legal position as the final move. Not only does this method fully utilize the thinking process, but it also effectively ensures the legality of the final move position. In practice, we consider the candidate move position and its first-order neighbors as positions to be evaluated. We first determine whether these positions are legal (if none are legal, we extend to second-order neighbors, and so on). Then, we use the LLM to score each legal position. Ultimately, the highest-scoring legal position is determined as the final move. The detailed process of this method is shown in Figure 4 and Figure 5.

## 3.4 Self-play

Although a wide range of Gomoku strategies and analytical logics were collected in Section 2.2, it should be noted that there are contradictions between these strategies and analytical logics, or some strategies are only applicable to specific game situations. Therefore, how to accurately select the most appropriate strategy and analytical logic based on the current game situation has become a crucial issue. Taking the chess-playing methods of machine learning models such as AlphaGo (Silver et al., 2016), AlphaGo-Zero (Silver et al., 2017) or AlphaZero (Silver et al., 2018) as an example, they usually use the Self-Play method, that is, the model plays games continuously with another instance of itself, and uses the final game results to improve the model's chess-playing level. In this process, the reward for the winner is set to 10, the reward for the loser is -10, and the reward for a draw is 0.

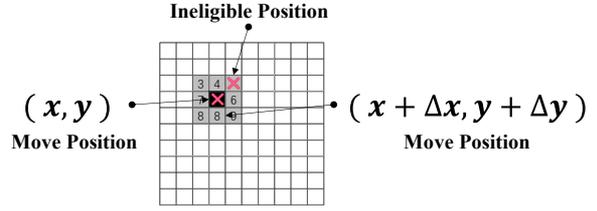

Figure 4: Schematic Diagram of Local Position

Drawing on this idea, our study adopts the method of self-play to exercise and improve the ability to select strategies and analytical logics in continuous practical combat, so as to achieve the goal of accurately selecting the optimal strategy and corresponding analytical logic for the current game situation. In the specific implementation, our study sets up two player Agents, representing black and white pieces respectively. Both use the same strategy selection model to choose strategies and analytical logics for their respective game situations, and continue to play alternately until the game ends.

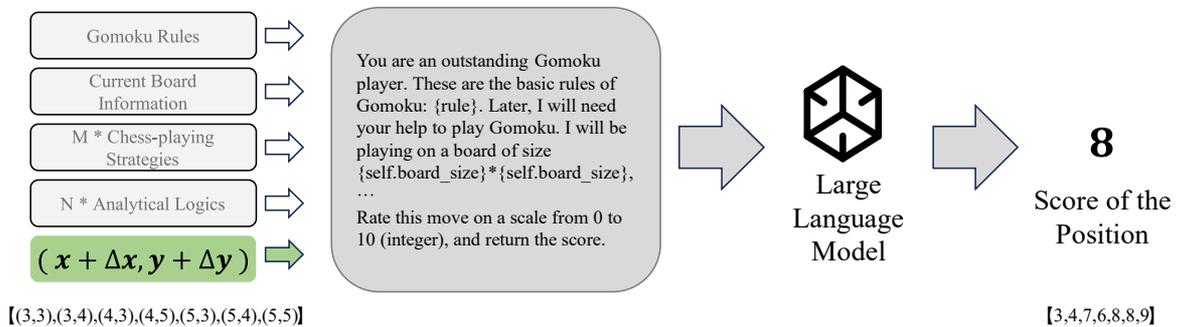

Figure 5: Flowchart of the Overall Process for Local Position Scoring.

## 3.5 Reinforcement Learning

Unlike conventional reinforcement learning models (Sutton and Barto, 1998), using language models for playing chess requires a significant amount of time to complete an entire game, from the beginning to the end. This greatly slows down the efficiency of improving the chess-playing ability. To address this issue, our study aims to introduce per-turn rewards in the middle of the game. Specifically, we designed a dedicated Agent for evaluating the game situation. This Agent assesses the two players in the current game, provides the win rate for each player, and uses these win rates as rewards to update the strategy selection model.

In this study, we employ a Deep Q-Network (Mnih et al., 2015) to conduct reinforcement learning training during continuous gameplay. The goal is to obtain the action values for each game situation, thereby selecting the optimal strategy and analytical logic. Specifically, s represents the board state before making a move, a represents the strategy action chosen by the Agent, r represents the reward obtained for that move, Q represents the action value function, and θ represents the network parameters. The detailed process is shown in Figure 4. The specific network structure consists of a three-layer MLP. The input size is 1515 (all positions on the board), and the output size is 529 (the combination of strategies and analytical logics).

## 4 Experiment

### 4.1 Parallel Position Evaluation

In Section 3.3, the local position evaluation requires scoring each legal position among the candidate move positions and their local neighbors one by one, which is usually very time-consuming, with an average speed of only "150 seconds per move," slowing down the reinforcement learning process. To tackle this challenge, we have carefully designed a parallel framework to achieve parallel scoring of all legal positions in all local areas. Specifically, we use Ray (Moritz et al., 2018) to build the parallel framework and assign a separate large language model to each position for position evaluation. These models operate independently and synchronously. Finally, we aggregate and sort the results output by each model and select the position with the highest score as the final move.

### 4.2 State-Action-Reward Database

In the process of conducting Gomoku research using language models, to enhance the model's chess-playing ability through multiple self-play sessions, it is necessary to frequently and concurrently call Deepseek-v2.5 API (DeepSeek-AI et al., 2024; DeepSeek-AI et al., 2025). However, in practice, due to the instability of API calls, the program often stops unexpectedly halfway, not only interrupting the self-play process but also causing the loss of all valuable information from the previously completed rollout process. This forces researchers to restart the program and begin the self-play from scratch, significantly reducing research efficiency and increasing research costs.

To address this thorny issue and avoid the loss of information due to mid-process interruptions, we have carefully designed and implemented an innovative architecture. The core functionality of this architecture is the ability to save and load "state-action-reward" pairs generated during all completed rollouts in real-time and accurately, systematically storing them in a specially constructed database. This database not only serves as an information repository, completely recording the key data of each self-play process, but also offers high flexibility and practicality for standalone model retraining. Through this approach, even if an unexpected interruption occurs during the API call, researchers can quickly restore the previous state from the database and seamlessly continue the self-play and model training, thereby greatly improving the continuity and efficiency of research and ensuring the stable progress of the research work.

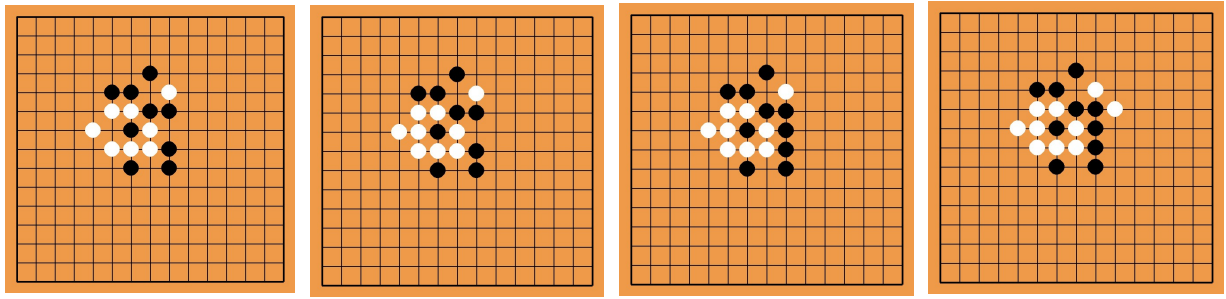

Figure 6: Visualization results of four consecutive moves in a particular game.

### 4.3 Visualization Module

To present the decision-making process of the language model in Gomoku games and its interaction with human players in a more intuitive, clear, and understandable manner, we have designed and added a dedicated visualization module. This module adopts a display format similar to that of the classic Gomoku game, vividly and graphically visualizing every detail of the gameplay and human-computer interaction. Through this module, observers can clearly see the position of each move, the strategy selection, and the dynamic changes in the situation, thereby gaining a deeper understanding of the decision-making logic and gameplay strategy of the language model. The specific visualization results are shown in Figure 6.

### 4.4 Results

Through the refined strategy and analytical logic selection mechanism, the language model has achieved significant success in selecting move positions in Gomoku. Compared with Zero-shot (Wang et al., 2019) , Few-shot (Brown et al., 2020), or direct Chain of Thought (Chu et al., 2023) methods, its performance has been notably improved. Moreover, the local position evaluation module introduced in this study has fundamentally solved the problem of the language model frequently outputting illegal positions during gameplay, which used to prevent the game from proceeding normally. This enables the language model to conduct self-play smoothly. In addition, by employing parallel position evaluation technology, the average speed of the entire process has been dramatically reduced from "150 seconds per move" to "28 seconds per move," achieving a leap of about 5 times in speed while ensuring stable performance. In the experimental part, this study fully utilized the powerful computing capability of 24 CPU cores and conducted 1,046 self-play games to train the Deep Q-Network. After such large-scale training, the model's Gomoku-playing ability has made a qualitative leap compared with the untrained state. Its decisions are more precise, and its gameplay level has been significantly enhanced, fully demonstrating the effectiveness and practicality of the methods in this study.

### 4.5 Performance Evaluation

To reasonably evaluate the performance of the language model in Gomoku as presented in this study, we employed two evaluation methods: human qualitative assessment and survival step evaluation against AlphaZero.

**Human Qualitative Assessment** This primarily relies on the subjective evaluation of human players regarding the model's chess strategies, decision-making rationality, and overall performance. This method can intuitively reflect the model's gameplay level from the perspective of human players.

**Quantitative Assessment** This is conducted by having the model play against AlphaZero (a model obtained from existing research) and recording the number of steps the model survives in each game to measure its performance. A total of eight games were played to ensure the reliability of the evaluation results. This quantitative method can objectively reflect the model's durability and competitiveness in actual gameplay.

Table1: Qualitative and Quantitative Evaluation Results of Different Methods.

| Methods | Smooth Game Completion Capability | Human-Assessed Playing Level | Average Survival Steps Against AlphaZero |
|---|---|---|---|
| Zero-shot | No | Very Poor | - |
| Few-shot(3) | No | Very Poor | 5 |
| Chain-of-thought | No | Poor | 6 |
| Random Strategy Selection | No | Poor | 7 |
| Local Position Scoring | Yes | Poor | 7 |
| 100 Episodes of Training | Yes | Poor | 9 |
| 500 Episodes of Training | Yes | Average | 11 |
| 1000 Episodes of Training | Yes | Average | 12 |

The detailed evaluation results are shown in Table 1.

## 5 Discussion

Although the method presented in this study has successfully enabled the use of language models to play Gomoku and trained a Gomoku player of a certain level through reinforcement learning, there are still some pressing issues that need to be addressed. First, the self-play process is too time-consuming, which makes it difficult for the model to quickly grasp some basic chess rules and requires a large number of games to gradually improve its playing ability. Second, in the selection of strategies and analytical logics, in order to simplify the reasoning process, this study only selects one strategy and one analytical logic for thinking each time, which to some extent limits the comprehensiveness and depth of the model's game analysis.

In future research, we plan to adopt a combination of multiple sets of "strategies + analytical logics" to more comprehensively evaluate and select the best chess strategies. In addition, we will explore the use of advanced deep reinforcement learning models such as Deep Deterministic Policy Gradient (Lillicrap et al., 2019), or the use of multi-agent systems (Albrecht et al., 2024), to further enhance the model's thinking ability in complex game situations. At the same time, we will also try to use the results of AlphaZero to guide the language model to think in the direction of the most correct move (Tie et al., 2025), thereby accelerating the speed of model capability improvement. In addition, we will also explore the use of state-of-the-art large vision-language models (Wang et al., 2025) to further enhance the model's performance.

## 6 Conclusion

This study successfully developed a Gomoku AI system based on LLMs. By enabling the model to understand the board and rules, learn strategies, analyze situations, select strategies, evaluate positions, and using self-play and reinforcement learning, the model can make decisions like human players. The system effectively resolves illegal move outputs, significantly enhancing decision-making speed and gameplay level. This study provides new insights for using LLMs in complex strategic games and expands their application possibilities in gaming. Future work will focus on optimizing the system through multi-strategy combinations, advanced reinforcement learning models, and leveraging AlphaZero results to achieve higher Gomoku playing ability.

**Acknowledgments**

This study was completed as the final project for the course on Deep Reinforcement Learning Methods (PKU-08403574) at Peking University. I would like to express my sincere gratitude to Professor Xinggong Zhang from Wangxuan Institute of Computer Technology, Peking Univeristy, for his guidance.